%% file: main.tex
\documentclass{IEEEcsmag}

\usepackage[colorlinks,urlcolor=blue,linkcolor=blue,citecolor=blue]{hyperref}
\expandafter\def\expandafter\UrlBreaks\expandafter{\UrlBreaks\do\/\do\*\do\-\do\~\do\'\do\"\do\-}
\usepackage{upmath,color}
\usepackage[super]{natbib}

\jvol{59}
\jnum{2}
\jmonth{February}
\jtitle{Computer}
\pubyear{2026}

\begin{document}

\sptitle{\footnotesize © 2026 IEEE.  Personal use of this material is permitted.  Permission from IEEE must be obtained for all other uses, in any current or future media, including reprinting/republishing this material for advertising or promotional purposes, creating new collective works, for resale or redistribution to servers or lists, or reuse of any copyrighted component of this work in other works. \\
This article has been accepted for publication in Computer. This is the author's version which has not been fully edited and content may change prior to final publication. Citation information: DOI 10.1109/MC.2025.3592765
}

\title{Beyond Speculation: Measuring the Growing Presence of LLM-Generated Texts in Multilingual Disinformation}

\author{Dominik Macko}
\affil{Kempelen Institute of Intelligent Technologies, Bratislava, Slovakia}

\author{Aashish Anantha Ramakrishnan}
\affil{The Pennsylvania State University, PA, USA}

\author{Jason S. Lucas}
\affil{The Pennsylvania State University, PA, USA}

\author{Robert Moro}
\affil{Kempelen Institute of Intelligent Technologies, Bratislava, Slovakia}

\author{Ivan Srba}
\affil{Kempelen Institute of Intelligent Technologies, Bratislava, Slovakia}

\author{Adaku Uchendu}
\affil{MIT Lincoln Laboratory, USA}

\author{Dongwon Lee}
\affil{The Pennsylvania State University, PA, USA}

\markboth{THEME/FEATURE/DEPARTMENT}{THEME/FEATURE/DEPARTMENT}

\begin{abstract}\looseness-1Increased sophistication of large language models (LLMs) and the consequent quality of generated multilingual text raises concerns about potential disinformation misuse. While humans struggle to distinguish LLM-generated content from human-written texts, the scholarly debate about their impact remains divided. Some argue that heightened fears are overblown due to natural ecosystem limitations, while others contend that specific ``longtail'' contexts face overlooked risks. Our study bridges this debate by providing the first empirical evidence of LLM presence in the latest real-world disinformation datasets, documenting the increase of machine-generated content following ChatGPT's release, and revealing crucial patterns across languages, platforms, and time periods.
\end{abstract}

\maketitle

\let\thefootnote\relax\footnotetext{DISTRIBUTION STATEMENT A. Approved for public release. Distribution is unlimited.\\
This material is based upon work supported by the Department of the Air Force under Air Force Contract No. FA8702-15-D-0001. Any opinions, findings, conclusions or recommendations expressed in this material are those of the author(s) and do not necessarily reflect the views of the Department of the Air Force.}

\chapteri{S}ince the widespread availability of LLMs through user-friendly interfaces like ChatGPT, concerns about their misuse have grown significantly \cite{zellers2019defending}. These concerns are supported by evidence of LLM adoption across scholarly papers \cite{haider2024gpt}, scientific peer reviews \cite{liang2024monitoring}, Wikipedia articles \cite{brooks-etal-2024-rise}, and content-driven social media \cite{sun2025aigeneratedtextworldalready}. While research demonstrates LLMs’ capability to generate disinformation \cite{zugecova-etal-2025-evaluation} and the computing field is exploring regulations and policies \cite{nahar2025generativeaipoliciesmicroscope}, the actual extent of real-world misuse remains unknown, limiting our understanding of the true impact. There are inconsistencies in research observations, some arguing that heightened fears are overblown due to natural ecosystem limitations \cite{simon2023misinformation}, while others identifying that specific ``longtail'' contexts face overlooked risks \cite{10702034}. Our study addresses this gap by providing the first comprehensive assessment of LLM-generated content prevalence in real-world disinformation contexts.

The capabilities of LLMs can be harnessed for various legitimate use cases (e.g., polishing or translation) to support diverse content production pipelines. However, due to their high scalability, low costs, and ease of use, LLMs can also be misused to promote disinformation in several ways. These range from generation of disinformation text in multiple languages, through text translation and polishing (particularly for non-native speakers), rephrasing (producing the same disinformation narrative in various forms), or targeting specific groups of audiences (e.g., personalization or micro-targeting).
The spread of LLM-generated disinformation can also lead to data corpora pollution, inducing factual errors and hallucinations in newer models trained on this data \cite{guo-etal-2024-curious}. Our analysis reveals increasing prevalence of LLM-generated content (reaching 2.15\% in 2024) in the MultiClaim dataset (containing professionally fact-checked multilingual social-media texts), confirming these are not merely speculated concerns. The societal implications are significant such that LLM-generated disinformation can erode democratic processes and societal trust and threaten online safety \cite{https://doi.org/10.1002/aaai.12188}.

This study is specifically focused on these research questions: \textbf{(RQ1)} To what extent are machine-generated texts present in existing disinformation datasets? \textbf{(RQ2)} Within labeled disinformation datasets, what is the distribution of machine-generated text in ``false'' and ``true'' content? \textbf{(RQ3)} How does the prevalence of machine-generated texts vary across disinformation datasets, languages, social media platforms, and time periods?

Our study makes several key contributions to understanding LLM-generated disinformation:
\begin{itemize}
\item[{\ieeeguilsinglright}] By validation on diverse datasets, our detection methods establish a robust analytical framework for examining real-world disinformation content, confirming both the increasing presence and prevalence of machine-generated texts in disinformation datasets over time.
\item[{\ieeeguilsinglright}] The distribution of LLM-generated content varies significantly across languages and platforms, revealing targeted patterns of misuse rather than uniform effects. This provides empirical validation for previously speculated concerns and unsupported fears about increased LLM deployment in disinformation campaigns.
\item[{\ieeeguilsinglright}] Most importantly, our findings underscore the urgent need for continued investigation and improved countermeasures, including enhanced detection methods and credibility assessment systems to preserve information integrity in our evolving digital landscape.
\end{itemize}

\section{DATASETS AND METHODS}

In this study, two groups of datasets are used, briefly described in Table~\ref{tab:datasets}. The first group (provided in the lower part of the table) is focused on evaluation of robustness to out-of-distribution data of the used machine-generated text detection methods, mostly containing existing multilingual labeled (the ground truth of machine-generated vs human-written texts) benchmark datasets. The second group of datasets (provided in the upper part of the table) contains real data from the wild, which we are using for estimation of the prevalence of machine-generated texts.

We are using two multilingual detectors, fine-tuned on different data for machine-generated text detection. Specifically, the Gemma-2-9b-it model has been selected for its best performance in the recent study \cite{macko2025increasingrobustnessfinetunedmultilingual}. It was fine-tuned for the binary classification task on 1) the train split of the GenAI dataset (the detector called \textit{Gemma\_GenAI}) and on 2) the train splits of MULTITuDE combined with MultiSocial datasets (the detector called \textit{Gemma\_MultiDomain}). For the efficient fine-tuning process, we utilized the QLoRA technique. Since the detectors are trained on different data, their detection capabilities complement each other.

The detectors are firstly evaluated on the existing multilingual datasets (test splits, not included in the detectors training). After showing the detectors are robust enough to be used in the wild (on out-of-distribution data), we use them on four existing datasets, presumably containing disinformation, misinformation or propaganda.

\begin{table*}[!t]
\vspace*{4pt}
\caption{Overview of the used datasets. The upper part of the table includes the datasets that we have used for estimating the prevalence of the machine-generated texts, and the lower part includes the specialized benchmark datasets for evaluating the detection performance of detectors.}
\label{tab:datasets}
\tablefont
\begin{tabular*}{38pc}{@{}p{50pt}p{400pt}<{\raggedright}@{}}
\toprule
Dataset &	Description \\
\colrule
MultiClaim \cite{moro_2025_15413169} &	A dataset of previously fact-checked claims by professional fact-checkers along with social-media posts gathered from the wild matching the claims. We have used data up to March 2024 (available when conducting this study). The final data covers a small portion of texts from prior to 2019 and thousands of samples for the years up to 2024. \\[3pt]
FakeNews \cite{Raza2024} &	A dataset of English news articles about US elections, collected from April to October 2023, containing GPT-4 annotations (reviewed by humans) considering the texts to be real or fake. \\[3pt]
USC\_X \cite{balasubramanian2024publicdatasettrackingsocial} &	A sampled portion of a multilingual US election Twitter/X dataset (usc-x-24-us-election), collected from May to November 2024 (the last tweets from November 6th, 2024). The pseudo-random sampling covers up to 10,000 tweets per each month. \\[3pt]
FIGNEWS \cite{zaghouani-etal-2024-fignews} &	A dataset of the multilingual shared task on news media narratives (covering bias and propaganda), including Facebook posts in 5 languages on Israel War on Gaza that were publicly posted between October 7, 2023, and January 31, 2024. \\[3pt]
\colrule
MULTITuDE \cite{macko-etal-2023-multitude} &	An official test split of a dataset of real news articles of various media sources in 11 languages, containing also the generated counterparts (based on their headlines) by 8 LLMs. \\[3pt]
MultiSocial \cite{macko-etal-2025-multisocial} &	An official test split of a dataset of real social-media texts in 22 languages from 5 platforms, containing also the generated counterparts (by using 3 iterations of paraphrasing) by 7 LLMs. \\[3pt]
SemEval \cite{wang-etal-2024-semeval-2024} &	An official test split of the multilingual track of the shared task 8 of SemEval-2024 workshop, containing student essays and news articles in 4 languages written by humans and generated by 7 LLMs. \\[3pt]
GenAI \cite{wang-etal-2025-genai} &	An official test split of the multilingual subtask of the COLING workshop of GenAI content detection task 1, containing texts of various domains of 29 sources in 15 languages written by humans and generated by 19 LLMs. \\[3pt]
MIX \cite{macko2025increasingrobustnessfinetunedmultilingual} &	A mixture of existing datasets on a binary machine-generated text detection task, covering various domains, 7 languages and over 75 generators. \\[3pt]
\botrule
\end{tabular*}\vspace*{8pt}
\end{table*}

Fine-tuned detectors are known to not generalize well to out-of-distribution data, requiring their classification thresholds to be carefully calibrated on expected data distributions. For measuring the prevalence of LLM-generated texts independently from distribution-specific calibrations, we use the Mean Score metric, which represents the probability (values from 0.0 to 1.0) of the text being generated by a machine (a higher value represents a higher probability).

Furthermore, we use a combined confident detection utilizing two fine-tuned detectors for the actual prediction of machine-generated texts. The prediction is positive if one of the detectors predicts the positive class with a probability of 1.0 and the other detector is not fully confident with the negative prediction (i.e., a positive class probability is > 0.0). Moreover, the Gemma\_GenAI confident positive prediction is disqualified (not taken into account) for languages that exceed a 0.1 false positive rate on existing labeled datasets, specifically for Arabic, German, Italian, and Russian. The other detector has not reached such a threshold of false positive rate for any language.

Table~\ref{tab:performance} shows the performance of such detection on existing multilingual benchmark datasets, containing human-written (the number of samples is provided in the Human column) as well as machine-generated (the number of samples is provided in the Machine column) samples. The table provides three performance metrics: False Positive Rate (FPR) is a ratio of how many human-written texts have been incorrectly predicted as positive, True Positive Rate (TPR, a measure equivalent to Recall of the positive class) is a ratio of how many of machine-generated samples have been correctly predicted as positive, and Precision reflects a ratio of how many of the predicted positive samples are actually machine-generated.

In regard to the evaluation using unlabeled data (where the truth is unknown), the key indicator is Precision, as it indicates the reliability of the identified positive samples. As showed in the table, we can assume that at least 93\% (the worst case) of the samples are correctly predicted as machine-generated texts, reaching up to 99\% on some datasets.

\begin{table*}[!t]
\vspace*{4pt}
\caption{Performance of the combined confident detection of machine-generated texts on the existing multilingual labeled datasets. The worst-case performance is boldfaced.}
\label{tab:performance}
\tablefont
\centering\begin{tabular*}{24.5pc}{@{}p{40pt}p{40pt}<{\raggedright}@{}p{40pt}<{\raggedright}@{}p{40pt}<{\raggedright}@{}p{60pt}<{\raggedright}@{}p{60pt}<{\raggedright}@{}}
\toprule
Dataset &	Precision &	FPR &	TPR &	|Human| &	|Machine| \\
\colrule
MULTITuDE &	0.9985 &	0.0093 &	0.7442 &	3236 (11\%) &	26059 (89\%) \\[3pt]
MultiSocial &	0.9969 &	0.0108 &	\bfseries 0.4988 &	17367 (13\%) &	121460 (87\%) \\[3pt]
SemEval &	0.9490 &	\bfseries 0.0579 &	0.9852 &	20238 (48\%) &	22140 (52\%) \\[3pt]
GenAI &	0.9955 &	0.0025 &	0.5194 &	73634 (49\%) &	77791 (51\%) \\[3pt]
MIX &	\bfseries 0.9358 &	0.0476 &	0.6916 &	99759 (50\%) &	100000 (50\%) \\[3pt]
\botrule
\end{tabular*}\vspace*{8pt}
\end{table*}

We aggregate the analyzed texts in the datasets for various aspects (time period, language, platforms, original dataset labels) to see how prevalence of machine-generated text differs across them. For the crucial era distinction in regard to massive usage of LLMs for text generation, we consider the release of ChatGPT to the general public in November 2022. For the sake of the pre-/post-ChatGPT release era, we count the whole year 2022 as a part of the pre-ChatGPT period.

\section{RESULTS}

To answer research questions of this study, described in the introduction, we first provide estimation of the prevalence of machine-generated texts in disinformation, then we follow by analysis of differences in this prevalence by various aspects, and finally, we provide estimation of prevalence of machine-generated texts in other existing related datasets.

\subsection{Evolution of Machine-Generated Disinformation over Time}

\textbf{The prevalence of machine-generated texts in the MultiClaim dataset of fact-checked multilingual social-media posts is increasing over time.}

\begin{figure}
\centerline{\includegraphics[width=18.5pc]{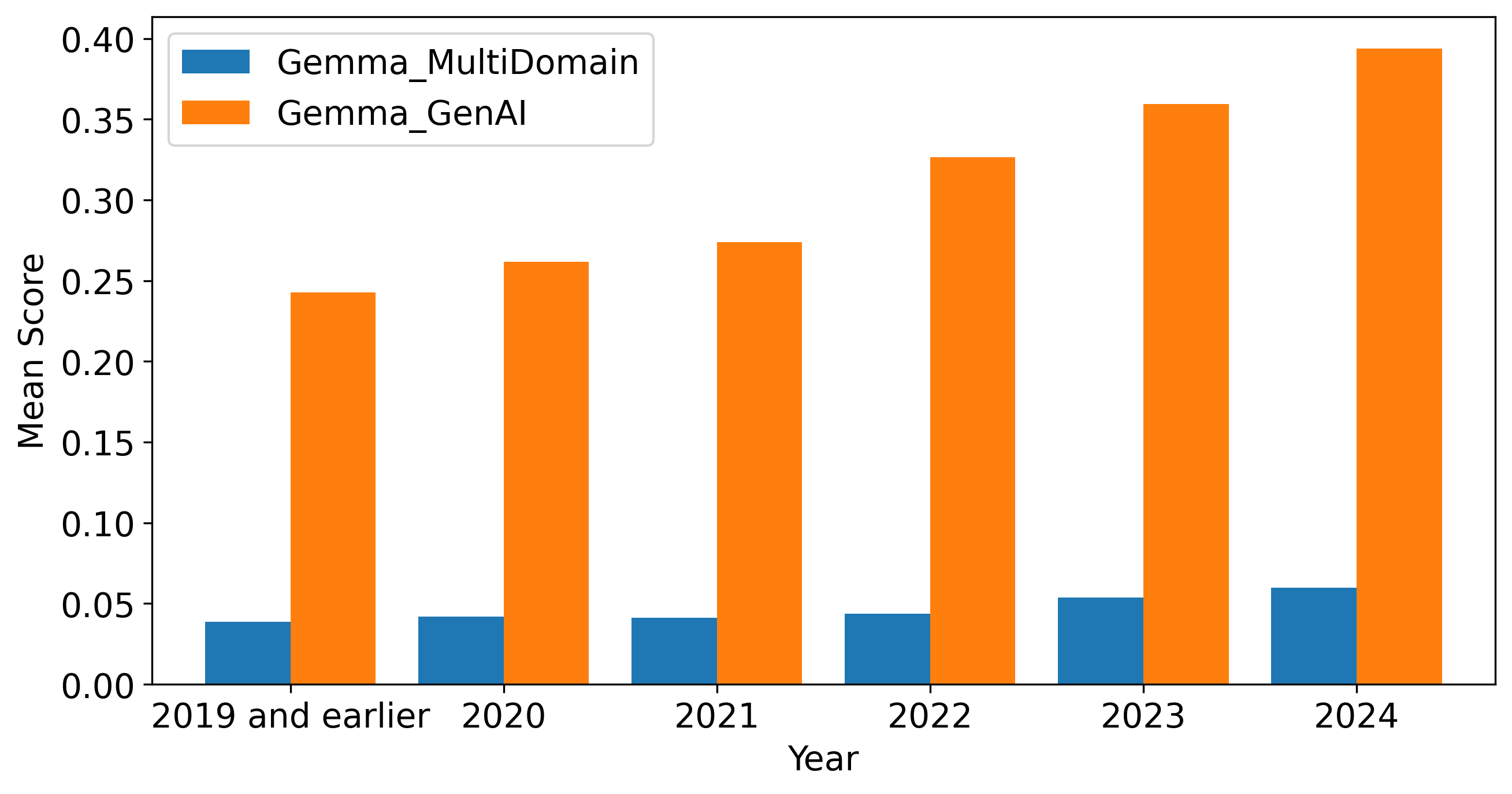}}
\caption{Per-year Mean Score of the two fine-tuned detectors for MultiClaim texts. Both detectors independently show increasing Mean Scores in recent years.}\vspace*{-5pt}
\label{fig1}
\end{figure}

One of the key objectives of this study is to evaluate whether the LLMs are already being misused for mis-/disinformation in the real world. Two multilingual detectors, fine-tuned on different data for machine-generated text detection (a binary classification task), show an increase of mean probability (Mean Score) of the texts grouped per year as being generated by the LLMs (Figure~\ref{fig1}). This provides evidence that in recent years (after the release of most-advanced popular LLMs, like ChatGPT), LLMs are increasingly being misused to generate content that is being professionally fact-checked. 

Although the years 2021-2023 contain approximately the same number of text samples (around 25,000), the Mean Score has increased from 0.04 and 0.27 in 2021 to 0.05 and 0.36 in 2023 for the Gemma\_MultiDomain and Gemma\_GenAI detectors, respectively. Such a 30\% increase of Mean Score roughly represents adding or changing of 1.3\% (for Gemma\_MultiDomain) to 13\% (for Gemma\_GenAI) of scores to 1.0 (i.e., maximal detector confidence). Since the detectors do not always provide a probability of 1.0 for each machine-generated text, the prevalence of LLM-generated texts between 2021 and 2023 has most probably increased even more. Based on this increase, we estimate that at least 1.5\% to 15\% of texts in 2023 of the MultiClaim dataset have been generated by LLMs.

\begin{figure}
\centerline{\includegraphics[width=18.5pc]{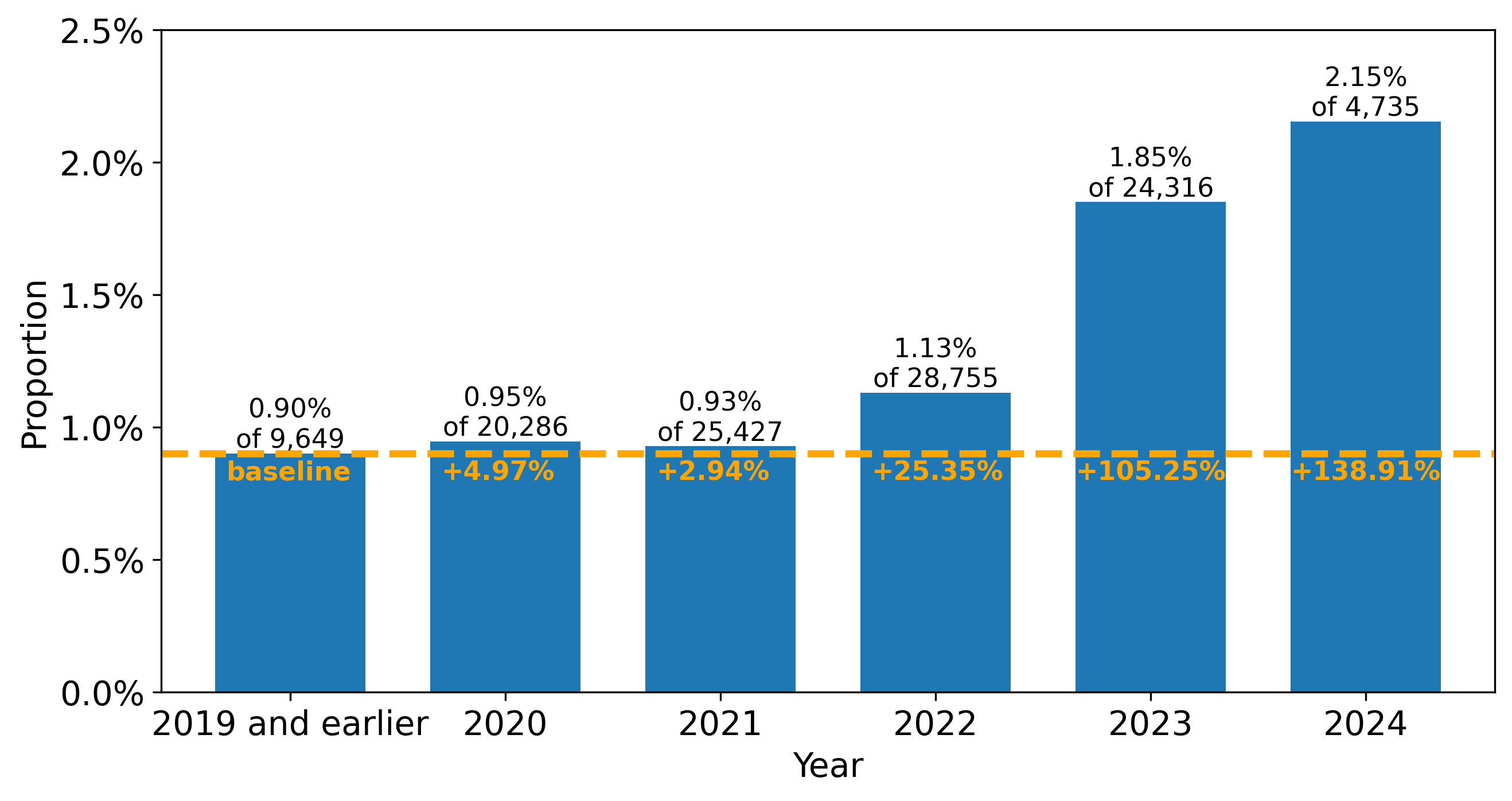}}
\caption{Per-year proportion of the MultiClaim texts detected to be machine-generated. Proportion is increasing in recent years, with the highest increase in 2023 (after ChatGPT release).}\vspace*{-5pt}
\label{fig2}
\end{figure}

To further analyze the proportion of the LLM-generated texts in the dataset, we combined their highest-confidence predictions (as described in the Methods section). Figure~\ref{fig2} illustrates the proportion of samples predicted (by highest certainty) to be generated by LLMs. The proportion has increased from 0.93\% in 2021 to 1.85\% in 2023 (i.e., a relative increase by 99\%). Given the worst-case calculated precision of the positive class (i.e., machine-generated) of such combined prediction on the existing multilingual labeled datasets (0.93), we can be highly confident that at least 1.7\% of texts in 2023 have been generated or modified by LLMs. The samples detected as ``generated'' in earlier years (especially 2019 and earlier) can form a baseline, representing either false positives or texts generated/modified by pre-LLM machines (e.g., translators).

\begin{figure*}
\centerline{\includegraphics[width=38pc]{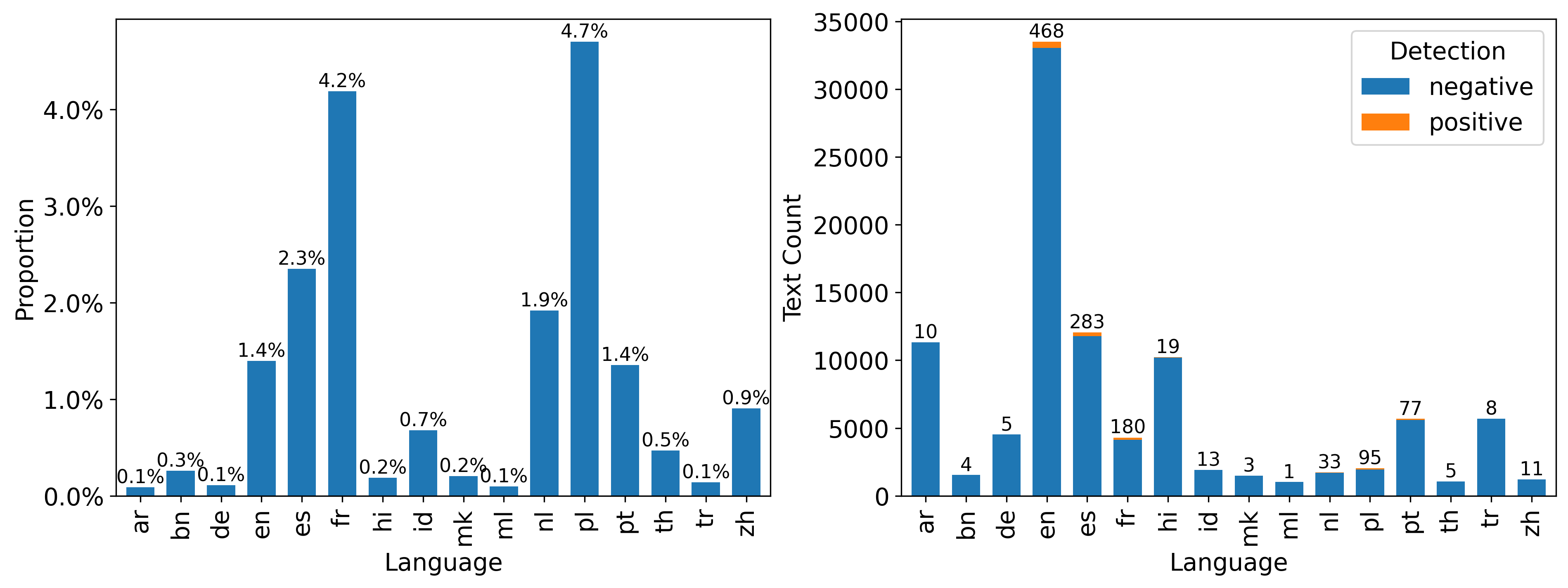}}
\caption{Left: Per-language proportion of the texts detected to be machine-generated (for languages including enough samples). Proportion differs among languages and is the highest in case of Polish and French. Right: Per-language number of samples showing also the distribution of samples classified as human-written (negative) and machine-generated (positive).}\vspace*{-5pt}
\label{fig3}
\end{figure*}

\subsection{Distribution of Machine-Generated Texts over Languages and Platforms}

\textbf{There are considerable differences in the prevalence of machine-generated texts in the MultiClaim dataset across languages as well as platforms.}

Analogously to the previous combined prediction, we have aggregated the results based on the languages, for which at least 1,000 samples are included in the dataset. The results are illustrated in Figure~\ref{fig3}, where although English and Spanish have the highest absolute number of texts predicted to be positive, the relative proportion shows to be the highest in the case of Polish (4.7\%) and French (4.2\%).

Regarding the contained social-media platforms, the differences are less severe, ranging from 0.64\% in Twitter (X), 1.29\% in Facebook, to 1.5\% in Instagram. Although Telegram contains over 1.7\% of LLM-generated texts, it covers only less than 1,000 samples in the dataset and, thus, is not objectively comparable to the others. Based on fact-checked claim verdict rating categories, the differences are not clearly observable, since ``False'', ``Misleading'', and ``Not categorized'' covered over 90\% of all samples. Among these three categories, the first one contains the highest proportion (1.36\%) of LLM-generated texts. Although the ``True'' category covers much less samples (786), out of them, 0.76\% of texts are identified as generated.

\subsection{Presence of Machine-Generated Texts in Other Datasets}

\textbf{There are machine-generated texts also present in other existing datasets and their prevalence differs among them.}

\begin{figure}
\centerline{\includegraphics[width=18.5pc]{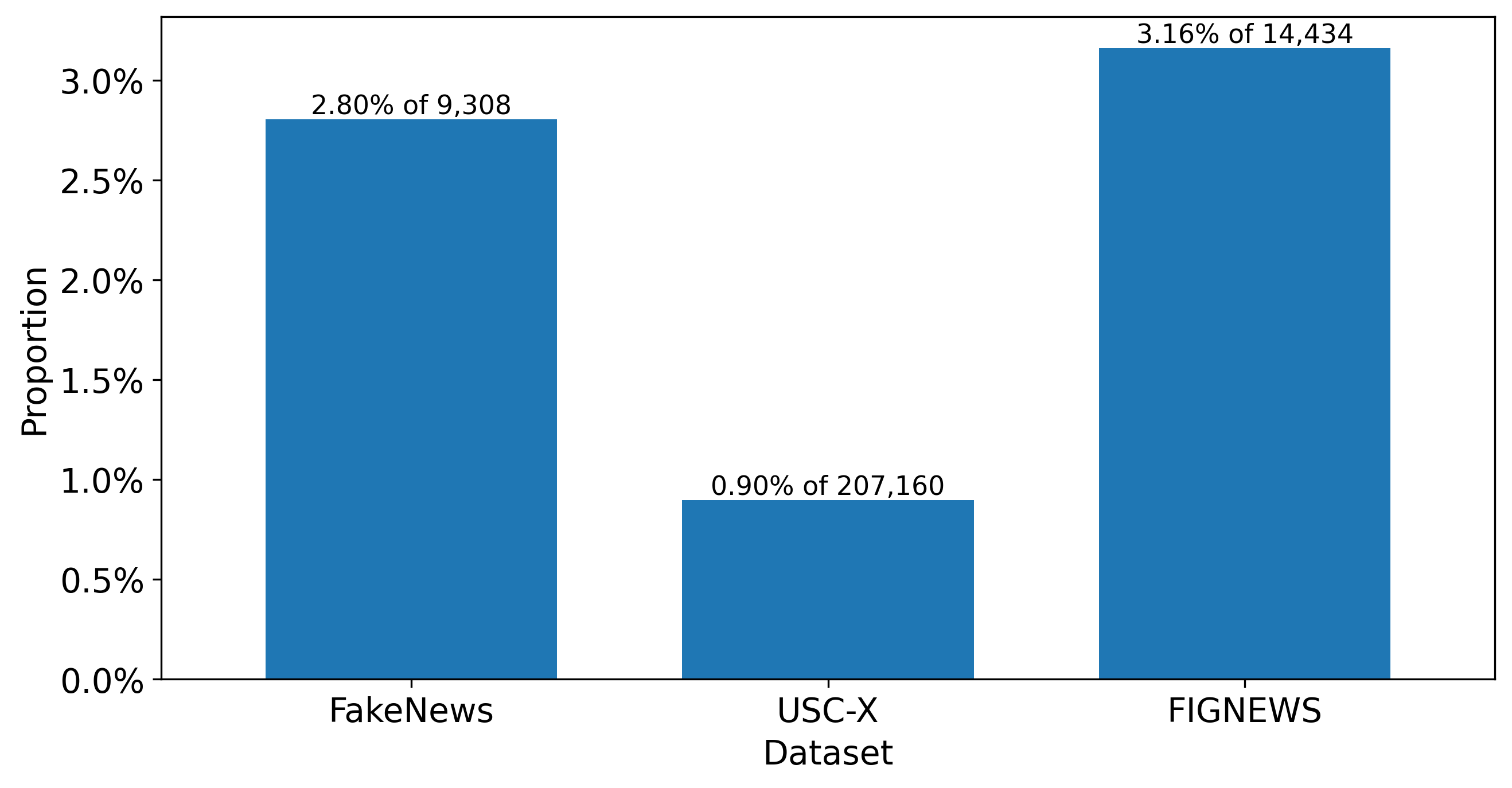}}
\caption{Proportion of the texts detected to be machine-generated in different datasets.}\vspace*{-5pt}
\label{fig4}
\end{figure}

We have similarly analyzed the presence of LLM-generated texts in other existing datasets, potentially including disinformation, misinformation or propaganda. The results illustrated in Figure~\ref{fig4} indicate that LLMs have generated or modified some of the texts also in these datasets. Although the proportions are not directly comparable to each other (due to containing a different number of samples of different kinds), FIGNEWS seems to contain the highest portion of machine-generated texts (3.16\%).

\begin{figure*}
\centerline{\includegraphics[width=38pc]{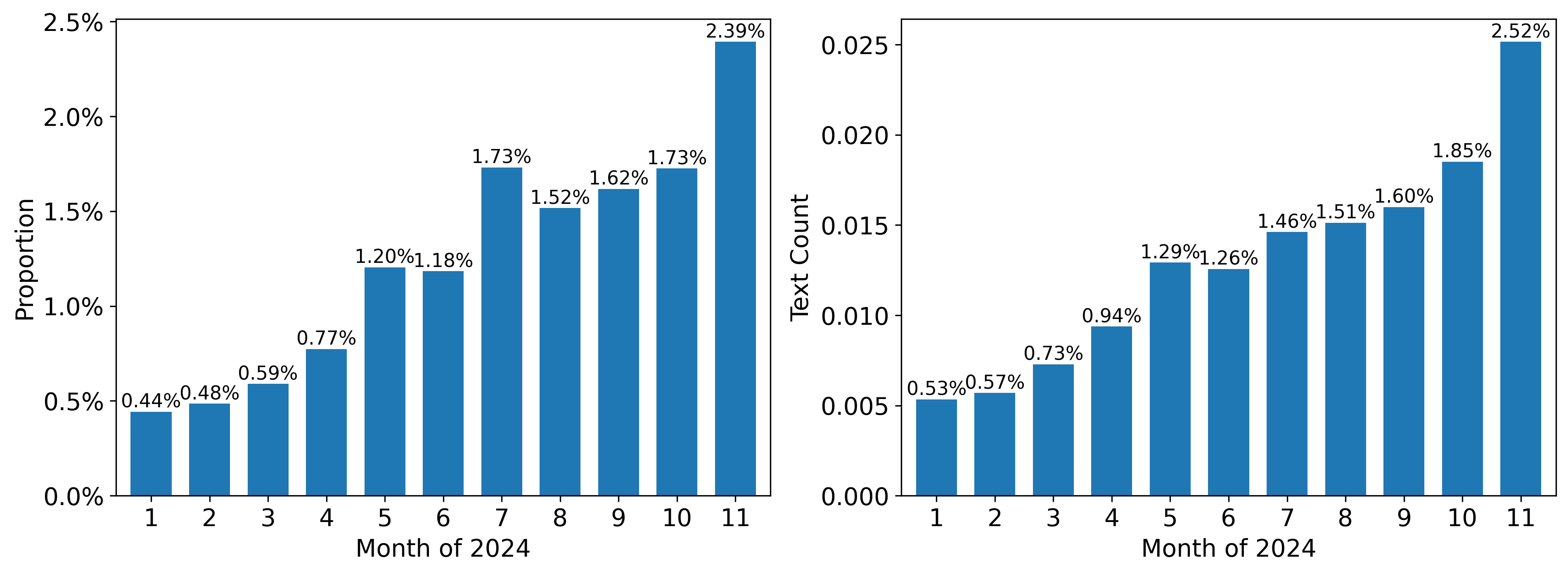}}
\caption{The per-month proportion of the texts detected to be machine-generated in the USC-X dataset for the election year of 2024 (all languages on the left, English texts only on the right). Proportion is increasing towards the election date in November 2024.}\vspace*{-5pt}
\label{fig5}
\end{figure*}

The FakeNews dataset contains the labels representing the text to be real or fake (labels verified by humans), including 2.61\% and 3.27\% of LLM-generated texts, respectively. It further provides evidence that LLMs are already being used for disinformation. Based on FIGNEWS dataset analysis, we found over 10\% of the French texts were being generated by LLMs, which is significantly more than around 4.7\% for English. On the other hand, the Hebrew, Hindi, and Arabic texts contained under 0.5\% of LLM-generated texts. We further found a higher prevalence of LLM-generated texts in the texts labeled as ``Biased against Israel'' (5.11\%) compared to the texts labeled as ``Biased against Palestine'' (3.13\%).

Although there is a rather low portion of LLM-generated texts in the USC-X dataset (0.9\%), it significantly differs across the months in the election year of 2024 (illustrated in Figure~\ref{fig5}). It has increased from 0.44\% in January to 2.39\% in November (i.e., an increase of 4.5 times). There is approximately the same number of samples in each month (8,700 to 10,000); thus, the proportions are reliably comparable.

Due to the space limitations, additional experimental results are provided in the Appendix.

\section{DISCUSSION}

Our findings reveal three key implications for understanding and addressing AI’s role in disinformation. First, \textbf{machine-generated content manifests unevenly across languages and platforms}, with some of them showing significantly higher prevalence. While broad ecosystem protections matter, targeted interventions (e.g., posts in a specific language of a specific platform) are needed in vulnerable contexts, particularly in languages such as Polish and French, where predicted prevalence rates reach 4.7\%.

Second, the \textbf{increasing temporal trend signals the importance of proactive monitoring}. The documented rise from 0.93\% in 2021 to 1.85\% in 2023 in the MultiClaim dataset \cite{moro_2025_15413169} demonstrates a clear trajectory requiring attention, particularly in high-prevalence contexts that may indicate broader trends. This suggests focusing early warning systems on languages and platforms showing higher concentrations of machine-generated content.

Although the detection methods can inherently have detection errors, such as false positives and false negatives, we have shown their capabilities on existing labeled multilingual benchmark datasets. The increasing prevalence in time corresponds with a common hypothesis that after ChatGPT’s release (November 2022), the prevalence of machine-generated texts has increased. Indeed, the most significant increase has been observed in 2023, when the proliferation of many modern LLMs with conversational interfaces enabled simplified usage for a wider community.

Third, our \textbf{empirical approach emphasizes the value of evidence-based assessment}. Rather than relying on speculation, our study provides concrete measurements of machine-generated content’s prevalence and distribution. It enables more targeted and effective responses to AI-generated disinformation.

\section{CONCLUSION}

Our results have demonstrated that LLM-generated texts are already present in the existing disinformation datasets coming from the real world, such as MultiClaim. Although the results do not indicate that LLMs are used for disinformation in significantly higher volume than for legitimate use, the fears of LLM misuse are justified. The MultiClaim dataset consists of fact-checked claims, but fact-checkers can only verify a small sample of the most viral or suspicious claims due to the vast daily influx of information. This creates a significant imbalance between true and false claims, complicating systematic comparisons.

The evidence of the presence of LLM-generated texts in datasets like USC-X or FIGNEWS may also imply that LLMs are already being misused for propaganda for elections or warfare. To enhance trust in the online information space, this further underscores the importance of indicating whether a given text has been generated or altered by an LLM. It would enable the readers to assess its credibility as such, considering that disinformation as well as unintentional misinformation can be generated by LLMs.

Future research should examine factors driving varying prevalence across contexts and investigate correlations with information ecosystem vulnerabilities. Maintaining and improving detection capabilities will be crucial as AI technology evolves, especially for vulnerable languages and platforms. Understanding these patterns and developing effective detection methods remains essential for preserving the integrity of information ecosystems.


\section{ACKNOWLEDGMENTS}
This work was partially supported by the European Union under the Horizon Europe project AI-CODE, GA No. \href{https://cordis.europa.eu/project/id/101135437}{101135437}; by the EU NextGenerationEU through the Recovery and Resilience Plan for Slovakia under the project No. 09I01-03-V04-00059; and by the Slovak Research and Development Agency under the project Modermed, GA No. APVV-22-0414. This work was also in part supported by U.S. National Science Foundation (NSF) awards \#2114824 and \#2131144. We acknowledge EuroHPC Joint Undertaking for awarding us access to Leonardo at CINECA, Italy.

\section{REFERENCES}
\def\refname{}

\bibliographystyle{IEEEtran}
\bibliography{bibliography}

\begin{IEEEbiography}{Dominik Macko}{\,} is a senior researcher at Kempelen Institute of Intelligent Technologies, Bratislava, Slovakia. 
Contact him at dominik.macko@kinit.sk.
\end{IEEEbiography}

\begin{IEEEbiography}{Aashish Anantha Ramakrishnan}{\,} is a Ph.D. candidate at The Pennsylvania State University, University Park, PA, 16802, USA. Contact him at aashish@psu.edu.
\end{IEEEbiography}

\begin{IEEEbiography}{Jason S. Lucas}{\,} is a Ph.D. candidate at The Pennsylvania State University, University Park, PA, 16802, USA. Contact him at jsl5710@psu.edu.
\end{IEEEbiography}

\begin{IEEEbiography}{Robert Moro}{\,} is a senior researcher at Kempelen Institute of Intelligent Technologies, Bratislava, Slovakia. Contact him at robert.moro@kinit.sk.
\end{IEEEbiography}

\begin{IEEEbiography}{Ivan Srba}{\,} is a senior researcher at Kempelen Institute of Intelligent Technologies, Bratislava, Slovakia. Contact him at ivan.srba@kinit.sk.
\end{IEEEbiography}

\begin{IEEEbiography}{Adaku Uchendu}{\,} is an AI Researcher at MIT Lincoln Laboratory, USA. Contact her at adaku.uchendu@ll.mit.edu.
\end{IEEEbiography}

\begin{IEEEbiography}{Dongwon Lee}{\,} is a professor at the Information School, The Pennsylvania State University, University Park, PA, 16802, USA. Contact him at dongwon@psu.edu.
\end{IEEEbiography}

\include{integrated_appendix}
\end{document}

%% file: integrated_appendix.tex
\expandafter\def\expandafter\UrlBreaks\expandafter{\UrlBreaks\do\/\do\*\do\-\do\~\do\'\do\"\do\-}



\setcounter{secnumdepth}{0}


\renewcommand\thefigure{A\arabic{figure}} 
\setcounter{figure}{0}

\section{APPENDIX}

\begin{figure}[!b]
\centerline{\includegraphics[width=18.5pc]{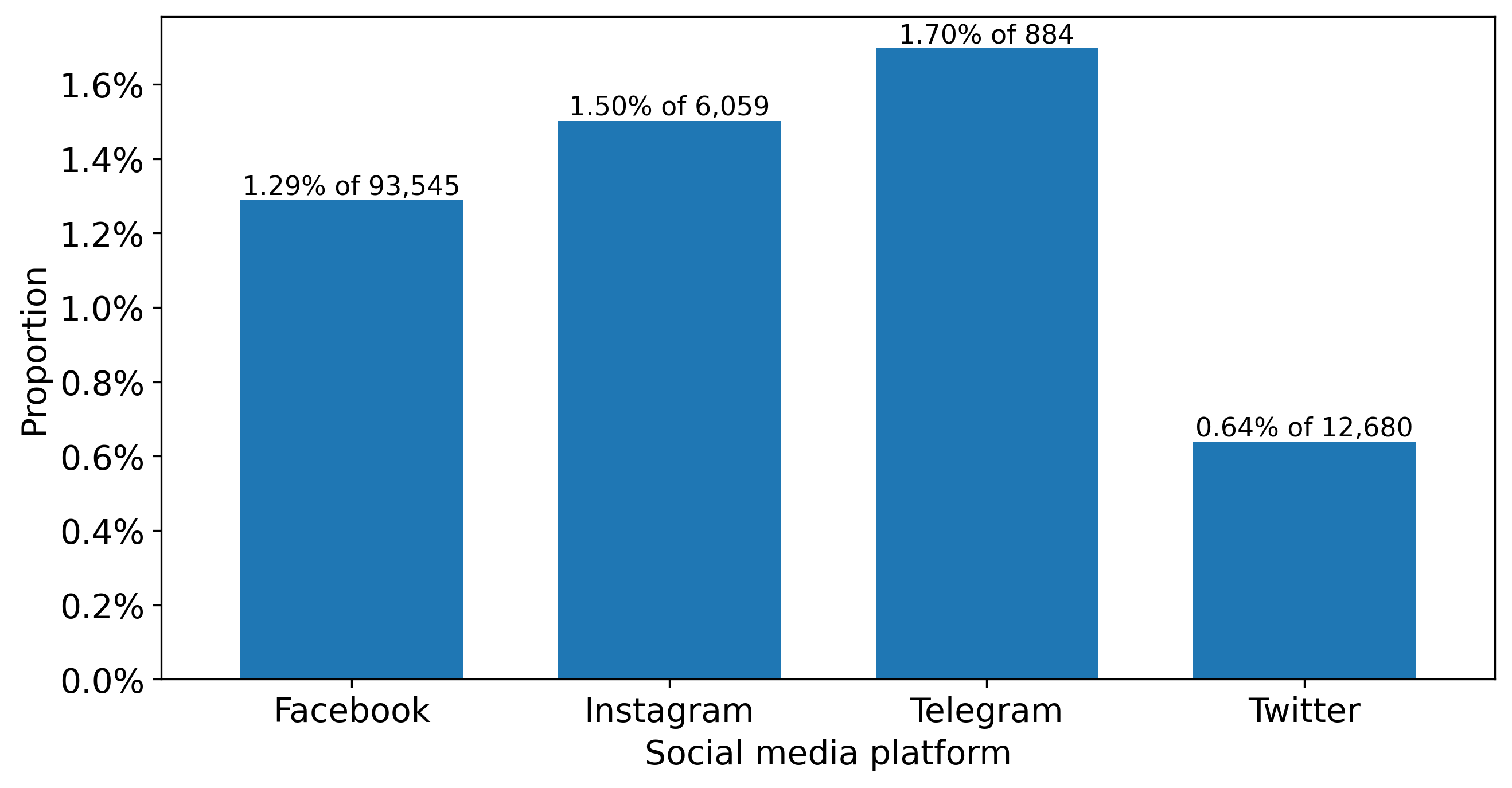}}
\caption{Per-platform proportion of the texts detected to be machine-generated (Telegram contains less than 1,000 samples). Proportion differs among platforms and is the lowest in Twitter (X).}\vspace*{-5pt}
\label{figA1}
\end{figure}

\begin{figure}[!b]
\centerline{\includegraphics[width=18.5pc]{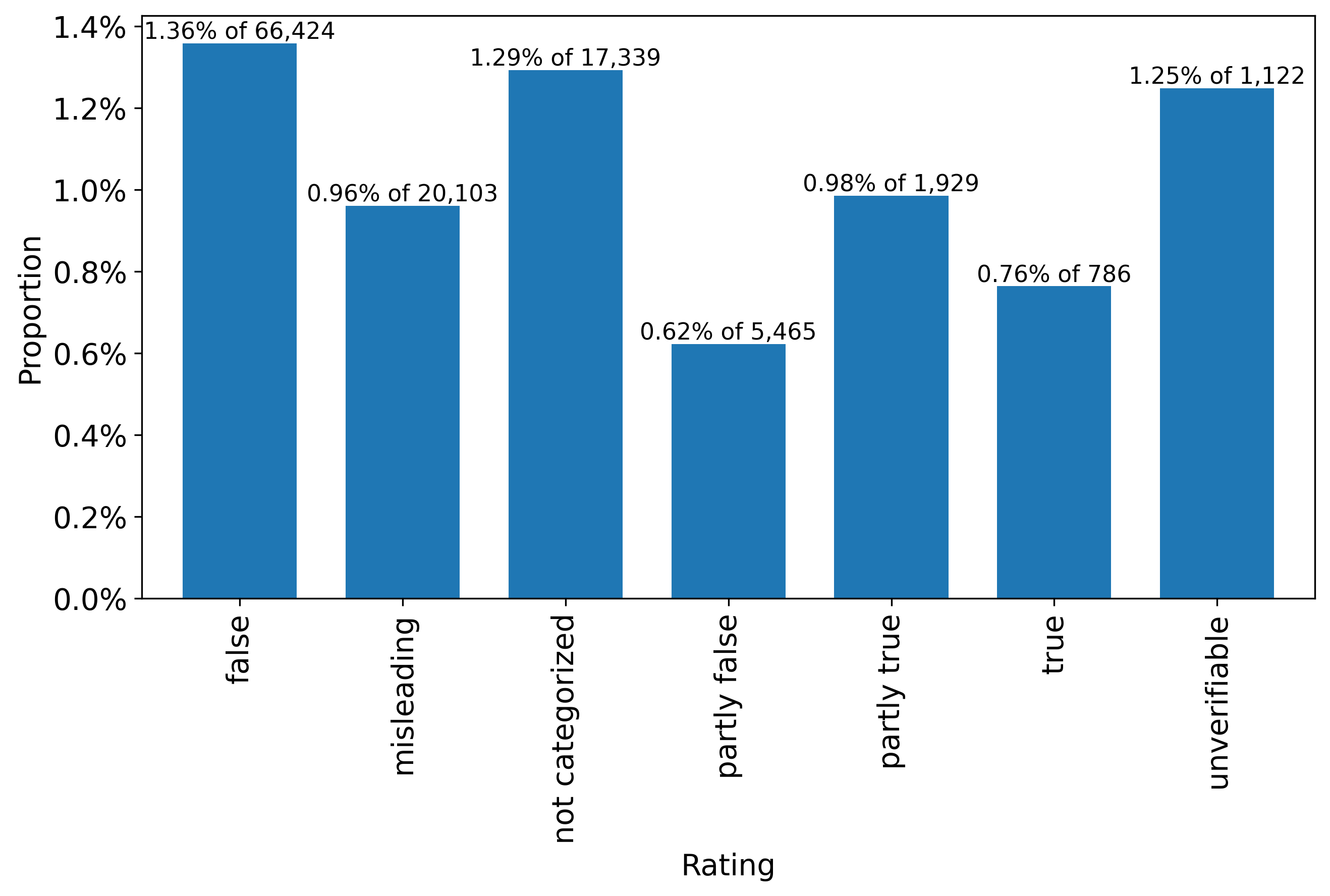}}
\caption{Per-rating proportion of the texts detected to be machine-generated. Proportion is the highest in case of ``false'' and ``not categorized'' texts. The ``true'' rating group covers less than 1,000 samples.}\vspace*{-5pt}
\label{figA2}
\end{figure}

\begin{figure}
\centerline{\includegraphics[width=18.5pc]{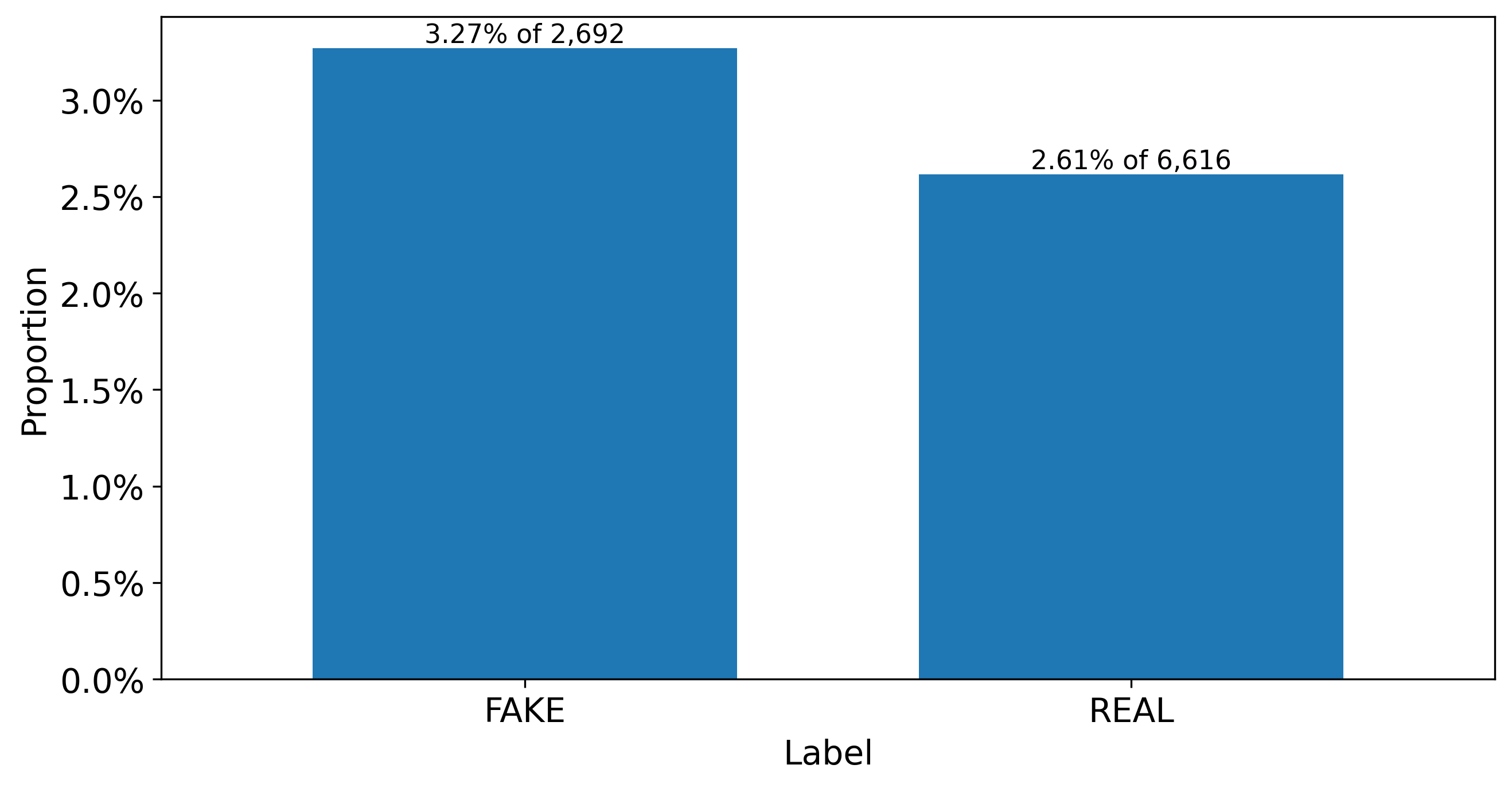}}
\caption{Per-label proportion of the texts detected to be machine-generated in the FakeNews dataset. Proportion is slightly higher in the ``FAKE'' texts.}\vspace*{-5pt}
\label{figA3}
\end{figure}

\begin{figure}
\centerline{\includegraphics[width=18.5pc]{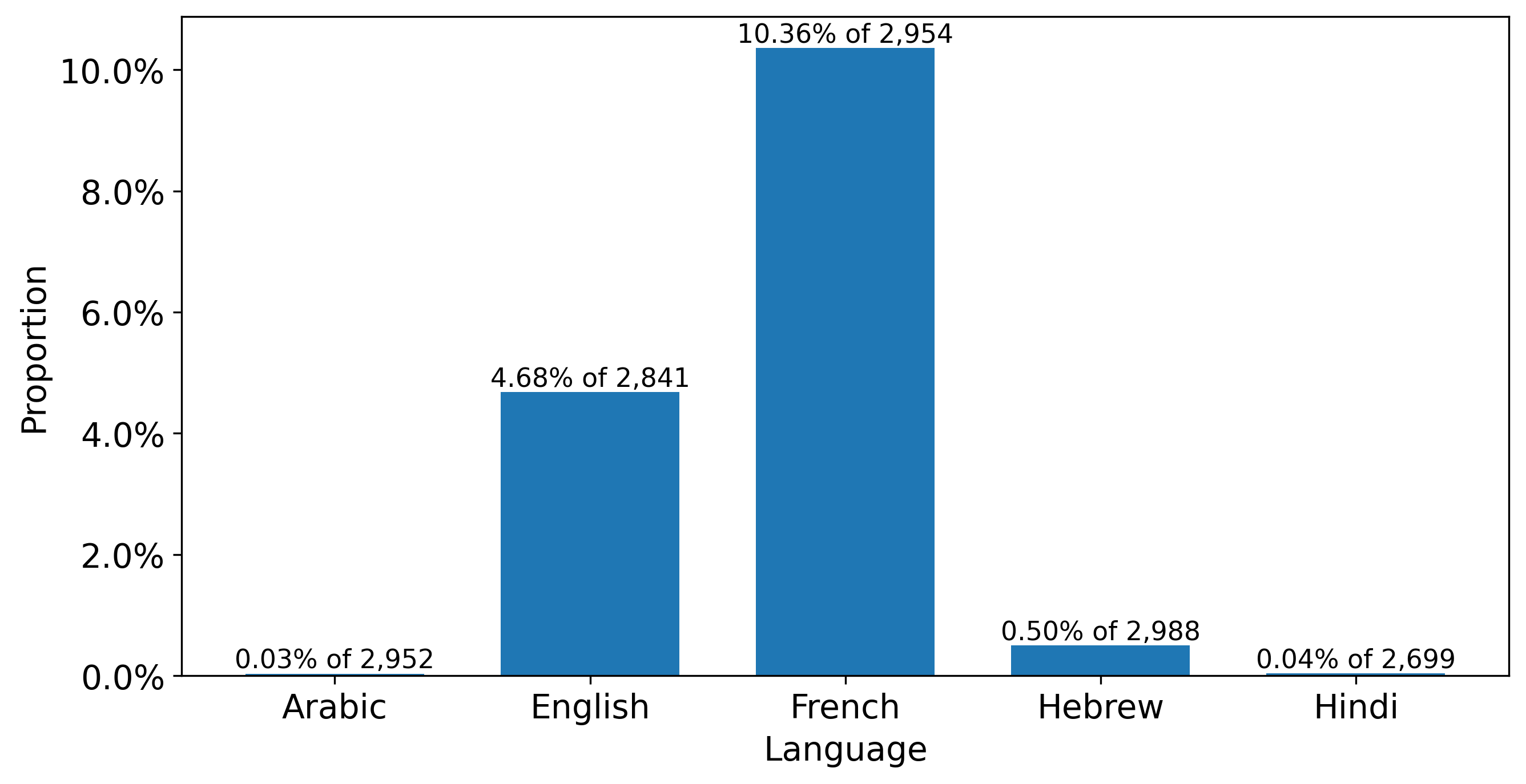}}
\caption{Per-language proportion of the texts detected to be machine-generated in the FIGNEWS dataset. Proportion is the highest in case of French and English.}\vspace*{-5pt}
\label{figA4}
\end{figure}

\begin{figure}
\centerline{\includegraphics[width=18.5pc]{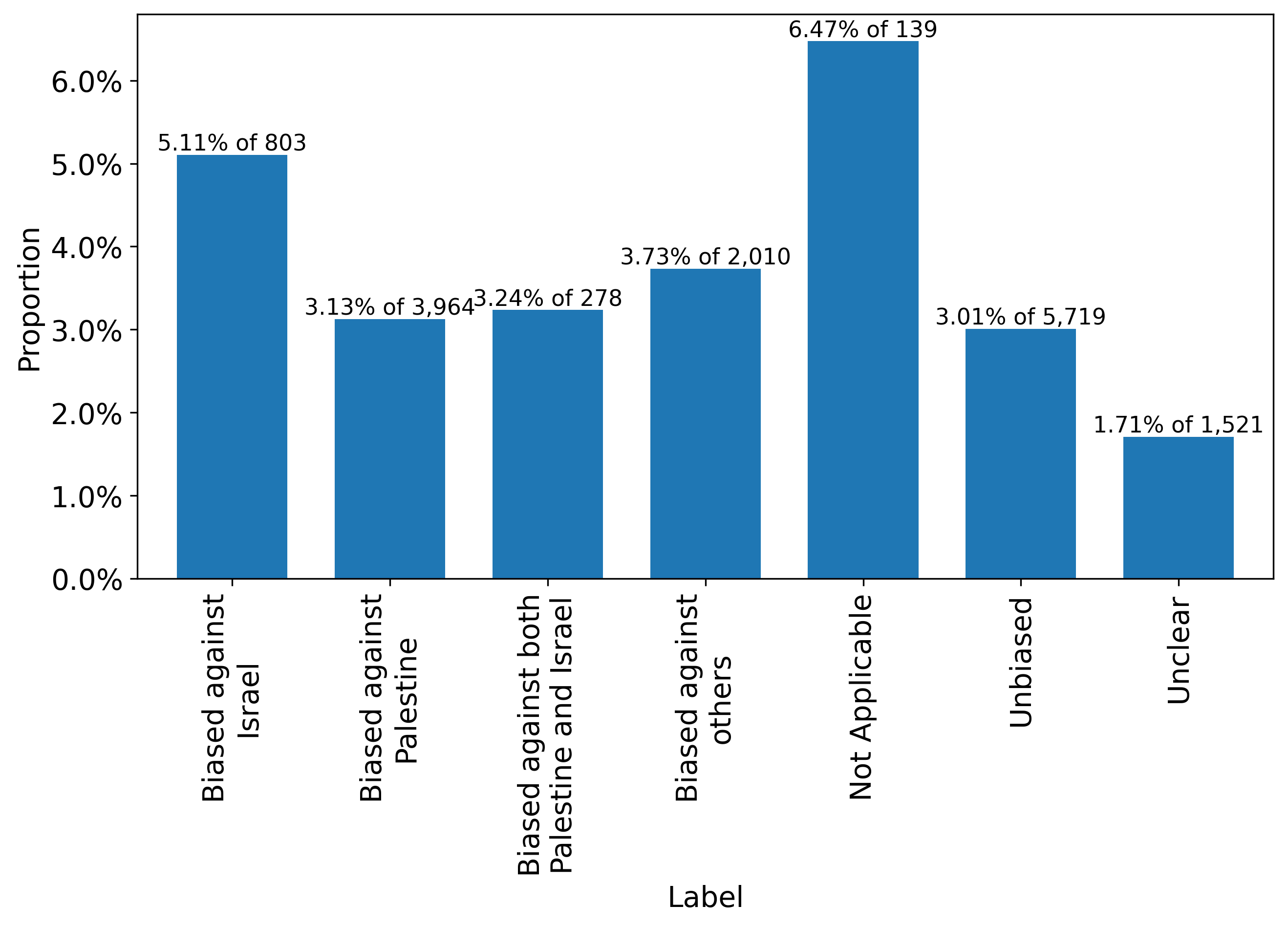}}
\caption{Per-label proportion of the texts detected to be machine-generated in the FIGNEWS dataset. Proportion is the highest for ``Not Applicable'' texts and for ``Biased against Israel'', both covering less than 1,000 samples.}\vspace*{-5pt}
\label{figA5}
\end{figure}

Figure~\ref{figA1} illustrates the proportion of machine-generated texts (as detected by the combined confident prediction) in the individual social-media platforms available in the MultiClaim dataset. Figure~\ref{figA2} illustrates the proportion of machine-generated texts in the MultiClaim dataset across the assigned fact-checking rating categories. Similarly, Figure~\ref{figA3}, Figure~\ref{figA4}, and Figure~\ref{figA5} illustrate the proportion of machine-generated texts in the FakeNews and FIGNEWS datasets based on available labels and language identification.

